\documentclass[conference, letterpaper]{IEEEtran}

\IEEEoverridecommandlockouts

\usepackage{multirow}
\usepackage{tabularx}
\usepackage{setspace}
\usepackage{pgfplots}
\usepackage{filecontents}
\usepackage{listings}
\usepackage{cite}
\usepackage{amsmath,amssymb,amsfonts, bm}
\usepackage{algorithmic}
\usepackage{graphicx,wrapfig}
\usepackage{graphicx}
\usepackage{textcomp}
\usepackage{xcolor}
\usepackage{balance}
\usepackage[T1]{fontenc}

\begin{document}

\title{FGLP: A Federated Fine-Grained Location Prediction System for Mobile Users}

\author{
\IEEEauthorblockN{
	Xiaopeng Jiang \IEEEauthorrefmark{1}
	Shuai Zhao \IEEEauthorrefmark{2}
	Guy Jacobson \IEEEauthorrefmark{3}
	Rittwik Jana \IEEEauthorrefmark{3}
	Wen-Ling Hsu \IEEEauthorrefmark{3}}
\IEEEauthorblockN{
	Manoop Talasila \IEEEauthorrefmark{3}
	Syed Anwar Aftab \IEEEauthorrefmark{3} 
	Yi Chen \IEEEauthorrefmark{2}
	Cristian Borcea \IEEEauthorrefmark{1}
}
\IEEEauthorblockA{
        \IEEEauthorrefmark{1}{
		Department of Computer Science, New Jersey Institute of Technology, Newark, NJ, USA}
}
\IEEEauthorblockA{
        \IEEEauthorrefmark{2}{
        Martin Tuchman School of Management, New Jersey Institute of Technology, Newark, NJ, USA}
        }
\IEEEauthorblockA{
		\IEEEauthorrefmark{3}{
		AT\&T Labs, Bedminster, NJ, USA}
\IEEEauthorblockA{		
		Email:\{xj8,sz255,yi.chen,borcea\}@njit.edu,\{guy,rjana,hsu,talasila,anwar\}@research.att.com}
}
}
\maketitle

\begin{abstract}
Fine-grained location prediction on smart phones can be used to improve app/system performance. Application scenarios include video quality adaptation as a function of the 5G network quality at predicted user locations, and augmented reality apps that speed up content rendering based on predicted user locations. Such use cases require prediction error in the same range as the GPS error, and no existing works on location prediction can achieve this level of accuracy. We present a system for fine-grained location prediction (FGLP) of mobile users, based on GPS traces collected on the phones. FGLP has two components: a federated learning framework and a prediction model. The framework runs on the phones of the users and also on a server that coordinates learning from all users in the system. FGLP represents the user location data as relative points in an abstract 2D space, which enables learning across different physical spaces. The model merges Bidirectional Long Short-Term Memory (BiLSTM) and Convolutional Neural Networks (CNN), where BiLSTM learns the speed and direction of the mobile users, and CNN learns information such as user movement preferences. FGLP uses federated learning to protect user privacy and reduce bandwidth consumption. Our experimental results, using a dataset with over 600,000 users, demonstrate that FGLP outperforms baseline models in terms of prediction accuracy. We also demonstrate that FGLP works well in conjunction with transfer learning, which enables model reusability. Finally, benchmark results on several types of Android phones demonstrate FGLP's feasibility in real life.
\end{abstract}


\section{Introduction}
\label{sec:intro}
{\fontfamily{ptm}\selectfont
A system that achieves high accuracy for fine-grained user location prediction on smart phones can be used by the OS and the apps to improve system or app performance~\cite{8570749}.
For example, since 5G performance is sensitive to small changes in location, the phone could use a map showing location-based quality of wireless network service to adapt video quality as a function of the predicted user locations. 
Augmented reality apps are delay-sensitive and can benefit from fine-grained location prediction to speed up content rendering. Yet another example is context-aware apps that need to adapt in advance based on where the user will move next, such as location-based gaming or advertising. 

Existing location prediction systems cannot be used in such scenarios. They work either at large spatial scales or at large time scales. For example, works for destination prediction~\cite{DeBrebisson:2015:ANN:3056172.3056178, 8758930, Hoch:2015:ELA:3056172.3056179,ijcai2017-430}, place-label prediction~\cite{7134083,7926813,8660411}, and Place of Interest (POI) prediction ~\cite{10.5555/3015812.3015841,10.1145/3178876.3186058,10.1145/3381006,8938733,9128016,ijcai2018-324,8290840,10.1145/2783258.2783350} have poor spatial accuracy (e.g., hundreds of meters). 
We are aware of one study predicting location at small time-scale (e.g., predict where the user will be in several minutes), but the location error is in the order of hundreds of meters~\cite{TRASARTI2017350}. There are additional works that focus on small time-scale check-in POI prediction ~\cite{xu2019venue2vec,zhang2019sparse}, but they do not work for fine-grained locations or for every location in a road network. 

The goal of this paper is to design a system for fine-grained location prediction that works on the users' phones and uses GPS traces. In our work, the term fine-grained refers to both spatial and temporal scales. Specifically, we aim to achieve high prediction accuracy, with prediction errors within the range of GPS errors.
We focus on pedestrians and bicyclists, instead of users in cars or in public transportation systems, because their less predictable behavior makes the problem more difficult. In addition, their lower speeds and ability to stop whenever they want are expected to enable more applications of predictions.
We also want to be able to predict at minute-scale (e.g., predict with a temporal step of one minute for $1, 2, ..., n$ minutes ahead). For example, we want to predict where the user will be in 5 minutes with a 10m error.

Our system would have limited usability if it could predict only places that have been visited previously by the user. The system can be improved by training the prediction model with data collected by all the users who adopt the system. While sharing location data across users will improve prediction accuracy, a naive method using GPS traces directly in centralized training is unlikely to be accepted by the users due to privacy concerns \cite{10.1007/s00779-008-0212-5}. 
Thus, the system needs to also provide location privacy protection. 
 
To summarize, we want to build a location prediction system that satisfies the following requirements: (i) achieves high prediction accuracy at a fine-grained spatio-temporal scale; (ii) works well for pedestrians and bicyclists;
(iii) works in places that have not been visited before by the owners of the smart phones invoking the prediction there; and (iv) protects user location privacy. To the best of our knowledge, there is no existing work that satisfies all these requirements.

This paper presents the Fine-Grained Location Prediction (FGLP) system that satisfies all the requirements mentioned above. FGLP has two main components: a federated, privacy-preserving learning framework and a prediction model. The learning framework runs on the smart phones of the users and on a server that coordinates learning from all users in the system. The framework represents the user location data as relative points in an abstract 2D space, which is a grid with fixed-size cells, that enables training on data received from all users. This representation scales all data to the same range and avoids the bias introduced by data with high longitude and latitude values, which weighs more during the deep learning optimization.

Our novel prediction model is trained on input derived from these relative points. The model uses Bidirectional Long Short-Term Memory (BiLSTM) and Convolutional Neural Networks (CNN), where BiLSTM learns the speed and direction of the mobile users, and CNN learns information such as user movement preferences. These two components are merged into a dense network with softmax activation. 
In FGLP, privacy is protected by combining federated learning (FL)~\cite{DBLP:journals/corr/abs-1902-01046} with our abstract 2D representation of user location data. FL trains the models locally on each phone and then computes a global model at the server by aggregating the gradients of the local models. In this way, the server never gets access to the data. However, the gradients of the local models may still leak private location information if the FL model uses physical location data~\cite{8737416}. This problem is substantially mitigated by our abstract 2D location data representation, which makes it difficult for the server to infer locations from model gradients.


Our experimental results, using a dataset with over 600,000 users, demonstrate that FGLP outperforms baseline models in terms of prediction accuracy for pedestrians and bicyclists.  We also demonstrate model reusability on another dataset, using FGLP with transfer learning~\cite{yosinski2014transferable}.
We benchmarked the model on three Android phones, 
and the results demonstrate that both training and inference are feasible in terms of execution time and battery consumption.

The rest of the paper is organized as follows. Section~\ref{sec:related} discusses related work.
Section~\ref{sec:framework} presents the federated, privacy-preserving learning framework. Section~\ref{sec:model} details the prediction model. Section~\ref{sec:data} describes the datasets and the data preprocessing. Section~\ref{sec:exp} shows the experimental results and analysis.
The paper concludes in Section~\ref{sec:conclusion}.

}


\section{Related Work}
\label{sec:related}
{\fontfamily{ptm}\selectfont
Early exploration of location prediction adapted Markov Chains and Hidden Markov Models~\cite{8570749}. Conventional machine learning (ML) methods, such as Bayesian networks~\cite{7134083}, Support Vector Machines (SVM)~\cite{7926813}, and tree-based models~\cite{Hoch:2015:ELA:3056172.3056179}, were also applied for location prediction. Due to the limited information extraction capability of these models, the performance suffered.

More recently, researchers have started to exploit deep learning (DL) techniques for location prediction by treating it as a time series prediction problem. In a taxi destination prediction competition, de Brébisson et al.~\cite{DeBrebisson:2015:ANN:3056172.3056178} tested several DL models, including MLP, LSTM, Bidirectional-RNN and Memory Network. Overall, the best model was Bidirectional-RNN with a time window covering 5 successive GPS points. In our case, given the need for fine-grained temporal scale, RNN-based methods alone cannot work well because they do not capture information such as road network characteristics and user preferences.
Another obstacle to directly adopting RNN-based methods is that the transition from one location to another cannot happen between any two locations~\cite{ijcai2017-430}. 
We overcome this by defining a reachable region centered at the current location and bounded by the traveling speed. 

The trajectory of movement on a map can be naturally processed with CNN-based methods. Lv et al.~\cite{8758930} proposed such a method, which beats the performance of Bidirectional-RNN~\cite{DeBrebisson:2015:ANN:3056172.3056178} in the taxi destination prediction problem. The method uses trajectory data to mark the visited cells in a grid-like space, but does not incorporate the visit frequencies at specific locations. The CNN component of FGLP, on the other hand, incorporates visit frequency, which helps to improve prediction accuracy. Zhang et al.~\cite{Zhang:2016:DPM:2996913.2997016} treated crowd inflow and outflow of grid cells in a city as a two-channel image-like matrix, and used CNNs for crowd flow prediction. This is a different problem from ours, but we share the ideas of visit frequencies for grid cells, and further extend the idea to represent the output as reachable grid cells.


Recent research ~\cite{10.5555/3015812.3015841,10.1145/3178876.3186058} applied state-of-art DL methods on POI IDs prediction. These works seem close to ours in terms of predicting human mobility. However, their problem definition is completely different, and their models use mechanisms that cannot work well for our problem. To predict the POI ID in a given time frame, these models formulate the problem as a recommender system that predicts a ranking of possible POI IDs a user will check in. On the other hand, our problem is formulated as a precise binary prediction of whether the user will be at a location cell over the grid or not. These existing models do not consider the spatial granularity, because a POI ID, such as a mall, typically covers a large area. They also do not consider temporal granularity, as they predict the next POI IDs without knowing when the user will visit that POI. Unlike these models, FGLP bounds the spatial and temporal granularity in the prediction and learns features such as speed and direction of travel, user preferences, and road characteristics. FGLP is unique in its fine-grained prediction across both spatial and temporal scales. 


None of the studies discussed so far attempted to provide location privacy.
Current privacy-preserving techniques in ML, such as differential privacy (DP), FL, and cryptographic methods could be applied for our problem, but have limitations. DP requires that computations be insensitive to changes in any particular individual’s record, thereby restricting data leaks through the results. However, recent studies show record-level DP fails to address information leakage attacks~\cite{10.1145/3133956.3134012}. A study by Graepel et al.~\cite{10.1007/978-3-642-37682-5_1} demonstrated machine learning on encrypted data using homomorphic encryption, but there are trade-offs regarding computational complexity and prediction accuracy. FL enables learning on the mobile devices without sending the raw data/features to the server. 
However, recent studies~\cite{8737416} showed user-level privacy leakage against FL by a malicious server, which can exploit the parameters received from the users. FGLP uses FL, but mitigates such attacks by using relative location, instead of physical location.

}
\section{FGLP Learning Framework}
\label{sec:framework}
{\fontfamily{ptm}\selectfont
\begin{figure}[t!]
  \centering
  \includegraphics[width=0.85\linewidth]{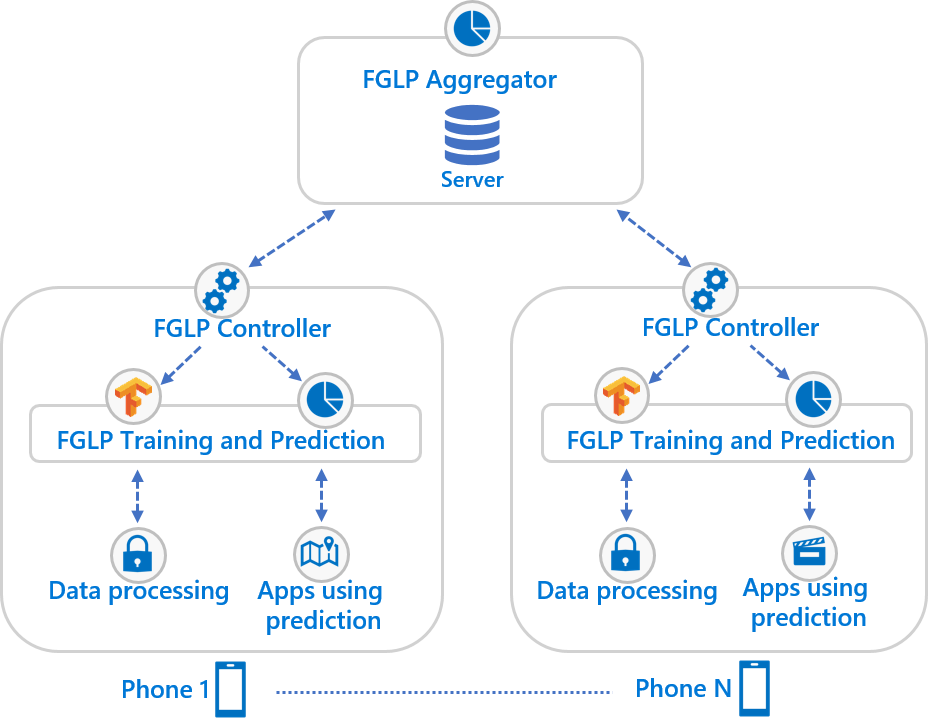}
  \caption{FGLP System Architecture}
  \label{FL-arch}
  \vspace{-0.15in}
\end{figure}

This section describes the FGLP framework that enables training and inference, while preserving the privacy of the user location. The section presents the system architecture for the framework, 
the operation stages of FGLP, and an enhanced training method for our federated system.

\subsection{System Architecture}

The system architecture of our framework is shown in Figure~\ref{FL-arch}. 
The framework software runs on a server and on the smart phones of the users, and it uses federated learning (FL)~\cite{DBLP:journals/corr/abs-1902-01046} for training across all users. The FGLP Controller on the phones mediates the communication between the server and phones. The Data Processing module on the phones processes the physical location data and generates the relative location data for training. The FGLP Training and Prediction module runs on the phones. This module performs local model training on the phones and then submits the model gradients to the server through the Controller. 
The FGLP Aggregator module at the server aggregates the gradients of the local models into a global model, and then distributes this model to the phones. When the OS or apps need a prediction, the Training and Prediction module is invoked. The output of the prediction is a relative location, which is then converted into a physical location, with help from the Data Processing module.

\subsection{Operation Stages}

To deploy the model in the real world and evolve it while the users collect location data over time, the FGLP learning framework has five computation and communication stages, as illustrated in Figure~\ref{FL}. During these stages, the smart phones and the server interact with each other and jointly contribute to the model. The stages are executed periodically in rounds, similar to Google's FL framework~\cite{DBLP:journals/corr/abs-1902-01046}. 
We also expect the performance to improve as more users adopt the system over time. In each round, the model is fine-tuned by re-training from the existing model. 
In the following, we detail each stage.

\begin{figure}[t]
  \centering
  \includegraphics[width=0.95\linewidth]{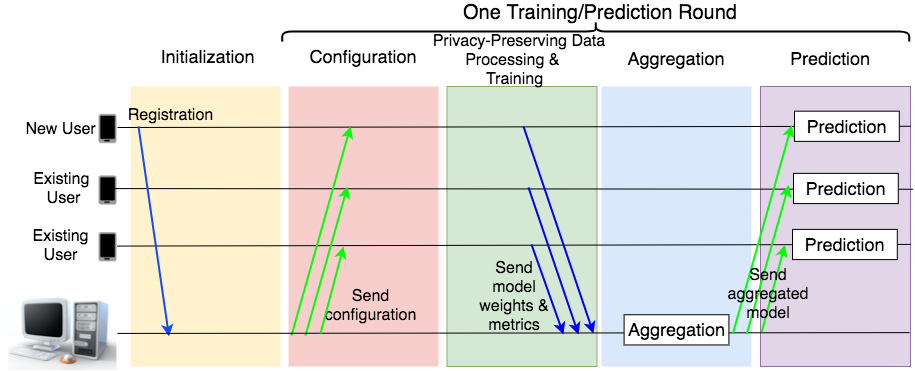}
  \caption{Federated Learning Operation of FGLP}
  \label{FL}
\end{figure}

(1)	\emph{Initialization.} 
Newly participating smart phones are required to register with the server in FGLP to ensure that the server knows when model gradients uploaded at different times come from the same user. It could further allow the server to remove potential malicious users who may inject fake data into the model.~\footnote{Protection against such malicious users is outside the scope of the paper.} 

(2)	\emph{Configuration.} A training/prediction round starts with the configuration stage. 
The server informs the smart phones of the deadline to participate in training. This deadline is the end of stage 3. 
The server can select a subset of the connected phones based on the optimal number of participating phones in each round and the availability of training data. 
The server sends configuration parameters to the phones on how to prepare the data for training. 
The server also sends the current global model parameters to each phone that did not participate in the previous training round, along with a training plan, such as gradient computation settings. 

(3) \emph{Privacy-Preserving Data Processing and Training.} Based on the configuration from the server, the phone performs first data processing. Then, it uses the global model received from the server to compute gradients based on its processed data. Finally, the phone sends the gradients back to the server after finishing the gradient computation.

(4) \emph{Aggregation.} 
The server waits for the phones to report gradient updates, aggregates them using federated averaging, and updates its global model weights with aggregated gradients. Then, it deploys the model to the phones to ensure they have the latest model because they may not participate in a new round for a while.

(5) \emph{Prediction.} The software on the phone can invoke the new FGLP model for predictions. Up to this stage, they use the old model from the previous round. 

\subsection{Training with Data Augmentation}
\label{sec:augment}

A well-reported issue to restrict the performance of models trained by FL is non Independent and Identically Distributed (IID) data distribution. FL trains on the dataset of each individual user. The datasets among different users may follow different distributions, because of user behavior differences, imbalanced class distribution, etc. While DL training can efficiently converge models with IID data, models trained in FL settings usually suffer from inferior performance~\cite{kairouz2019advances}. To mitigate the non-IID issue, we leverage an advanced data augmentation mechanism inspired by Zhao et al.~\cite{zhao2018federated}. 

In this enhanced FL training, the data from a small percentage of users (e.g., less than 5\%) are allocated as an augmentation dataset and made available to the aggregation server, and the server can sample and share it with the other users. This is usually the case when a small amount of users are willing to share their data with the server~\cite{zhaoprivacy}. Let us note that the users who perform on-device training and testing do not share any data with the aggregation server. 

Training with data augmentation has three phases. First, the FGLP model is trained with the augmentation dataset at the server. This model is then distributed to the phones that will participate in FL training. Second, each phone selected in every round randomly picks a certain number of samples from the augmentation dataset and concatenates them with its own dataset. Third, on-device training is conducted by initializing the model received from the server (trained with the augmentation dataset) and further training with the augmented local data. The rest of the FL procedures are the same as in the basic FL technique. Thus, the local non-IID data of each user are augmented with IID data from the augmentation dataset, and the non-IID issue is mitigated. 

}
\section{FGLP Model}
\label{sec:model}
{\fontfamily{ptm}\selectfont
This section presents our novel abstract data representation, the formal problem definition for our model, and the description of the model architecture and its components. 

\subsection{Abstract Data Representation}
 
The fundamental information to predict location is travel direction and speed. The user movement preferences and road characteristics also help the prediction. The GPS trajectories of each user contain this information. FGLP on the phones process the raw GPS data to generate relative trajectories. The Data Processing module in our framework represents trajectories as relative points in an abstract 2D space. 
 
\subsubsection{Raw Data}
The raw data is recorded by each phone using the embedded GPS sensor. Let $\mathbf{L_t = <lat_t, lon_t>}$ denote the latitude and longitude of a user at time t. FGLP performs learning based on the transportation mode, such as walking or bicycling. Therefore, only the data specific to the desired transportation mode is selected for further processing.
 
 \subsubsection{Input}
The raw data of each user is processed on their smart phone to produce two types of inputs: {\it fixed-length sequences of relative points} and {\it historic region occupancy matrices} of the space considered for prediction. The input sequences contain the speed and direction information of the user trajectories. The occupancy matrices record frequently visited places and the most likely trajectories between these places. The inputs are computed offline 
and can be updated over time based on new data to enable re-training.
 
To generate the input sequences, FGLP splits the user trajectories into fixed-length sub-trajectories. The length in time of the trajectories is determined experimentally. Each sub-trajectory is transformed into a sequence of relative points in an abstract 2D space.
The X and Y coordinates of relative points at time t are determined based on their offsets from the location at previous time step t-1. The location of the very first point in a trajectory session is excluded. A location offset is denoted as $\mathbf{\Delta L_t = <lat_t-}$ $\mathbf{lat_{t-1}, lon_t-lon_{t-1}>}$. An input sequence of at time t that looks back k steps is denoted as $\mathbf{S_t = (\Delta L_{t-k+1},}$ $\mathbf{\Delta L_{t-k+2},...,\Delta L_{t-1},\Delta L_{t})}$. In its training, FGLP considers all possible k-length sequences, including overlapping sequences.

The historic region occupancy matrices are extracted from a historic occupancy matrix of the entire space (e.g., a city). FGLP divides the entire space into a grid of fixed-size cells, and each cell corresponds to an element in the historic occupancy matrix. Each element represents the number of visits of the user in its corresponding cell. The matrix represents the occupancy of a bounded region $\mathbf{R_t}$ with area $A$, which is centered at the physical location $\mathbf{L_t}$ at time t. $\mathbf{R_t}$ is divided into $M \times M$ fixed-size grid-cells, where $A$ and $M$ are predefined constants based on the maximum speed of users and the desired spatial granularity for the prediction. Each historic region occupancy matrix $\mathbf{H_t}$ is a $M \times M$ matrix, and it is extracted from the historic occupancy matrix for the entire space. Once extracted, this matrix is an abstract input that does not maintain any relation with the physical locations that it represents. A matrix can implicitly tell if a road exists in a given cell (i.e., non-zero value for the corresponding matrix element) and can also tell if adjacent cells form routes taken frequently by the user. 

\begin{figure}[t!]
  \centering
  \includegraphics[width=1\linewidth]{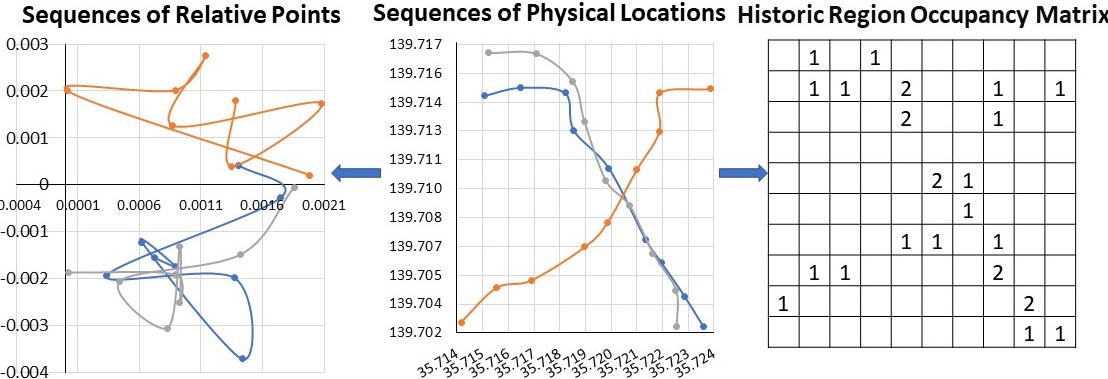}
  \caption{Illustration of Abstract Data Representation}
  \label{data_abstraction}
  \vspace{-0.15in}
\end{figure}

Figure~\ref{data_abstraction} illustrates how the input is created from the sequences of physical locations. We observe that the sequences of relative points do not resemble the physical sequences, which helps with location privacy protection. Furthermore, different physical locations can be mapped to the same relative locations. Similarly, the occupancy matrices of different areas may also be same.
 
\subsubsection{Output}
The location to be predicted $\mathbf{L_{t+i}}$ is mapped into the region $\mathbf{R}$.
FGLP builds a prediction matrix  $\mathbf{Y_{t+i}}$ (equation~\ref{eq:output}).

 \begin{equation}
 \label{eq:output}
\mathbf{y_{i,j,t+i}} = 
    \begin{cases}
        1, & if\;\mathbf{L_{t+i} \in R_{i,j}} \\
        0, & otherwise
    \end{cases}
\end{equation}

where $\mathbf{y_{i,j,t+i}}$ is an element of $\mathbf{Y_{t+i}}$, $\mathbf{R_{i,j}}$ $\mathbf{(1 \le i, j \le M)}$ is a cell in region $\mathbf{R}$. 
The output is formulated as a categorical class rather than a numerical value, so that we can set the spatial granularity of the prediction as a constant. Another reason for using categories is that the historic region occupancy matrix does not contain information to predict with spatial granularity beyond the grid-cell size. Overall, the output is a relative grid-cell, which is translated into a physical grid-cell on the user's phone.

\subsubsection{Benefits}
\label{sec:benefits}

This data representation has three benefits. First, different relative locations have the same magnitude, which is required for DL data. DL algorithms minimize the distance between two data points as a loss function. During this minimization, high-magnitude data weights more than low-magnitude data, and it can lead to bias. For example, if physical location sequences are used directly, the training will focus on minimizing the loss for the data with high latitude and high longitude values. One way to avoid this problem is to scale every sequence to the same range ~\cite{bishop1995neural}. However, scaling the location sequence will remove the traveling speed, which is necessary for location prediction. Since speed cannot be assumed constant for accurate prediction, there may not be an efficient mechanism to preserve it. With our data representation, all relative locations are in the same range and can be used directly in DL. This is especially important for FL training across all users.

The second benefit is the ability to change configuration parameters in data representation to perform prediction at different spatial granularity. For example, the grid-cell size can be 10m $\times$ 10m for pedestrians, and 40m $\times$ 40m for bicyclists. 

The third benefit is location privacy protection. This is achieved in conjunction with FL, which shares only the model gradients with the server. The input does not leave the phones because the learning happens on the phones. However, the gradients of the local models may still leak private location information if the FL model uses physical location data~\cite{8737416}. This problem is substantially mitigated by our abstract 2D data representation. The input contains the essential information for location prediction, including speed, direction, and user movement preference, while not disclosing the physical location ($\mathbf{L_t = <lat_t, lon_t>}$) of the user to DL model. 

\subsection{Problem Definition}

Let $\mathbf{S_t} \in \mathbf{R}^{2k}$ be the size-$k$ sequence of relative points at time $t$ for a given user. Let  $\mathbf{H_t} \in \mathbf{Z^+}^{M\times M}$ be the historic regional occupancy matrix of the same user, which is a square matrix of order $M$ centered at the user location at time $t$. Our goal is to predict the relative location of this user $\mathbf{\hat{Y}_{t+i}} \in \mathbf{Z_2}^{M\times M}$ for the future $i^{th}$ timestamp.

\begin{equation}
\mathbf{\hat{Y}_{t+i} = F(S_t; H_t)}
\end{equation}

where $\mathbf{F}$ is our DL model for location prediction. The predicted location is a cell in the $M \times M$ grid representation of the space surrounding the current location of the user.

\subsection{Model Architecture}

\begin{figure}[t!] 
  \centering
  \includegraphics[width=\linewidth]{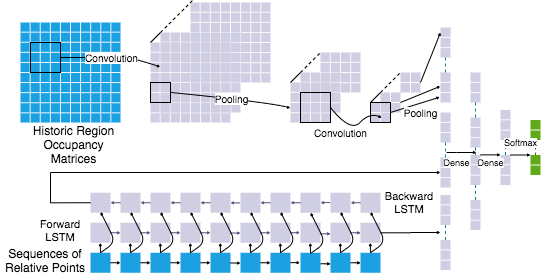}
  \caption{Model Architecture}
  \label{architecture}
\end{figure} 

FGLP can extract features essential to location prediction without using the physical location. As shown in Figure~\ref{architecture}, the model fuses BiLSTM and CNN, where BiLSTM learns the speed and direction of the user mobility from the sequences of relative points, and CNN learns user movement preferences and likely user routes from the historic region occupancy matrix. BiLSTM and CNN work in parallel. A densenet-type connection is used to fuse BiLSTM and CNN, and then softmax activation is adapted to output which grid-cell the user will be in. Batch normalization and dropout layers are also added in both BiLSTM and CNN to avoid over-fitting, but for simplicity they are not shown in the figure. This architecture is designed to capture as much user-level information as possible.

For training, the phones use the input derived from the relative data in the abstract space, which can be pre-computed.
For prediction, the sequence input is simple and can be generated in real-time, based on the last $k$ recorded GPS locations. The historic region occupancy matrix, centered at the current location, is extracted in real-time from the pre-computed historic occupancy matrix for the entire space. 

\textbf{BiLSTM.} 
FGLP uses BiLSTM to train on sequences of relative points. An LSTM unit is composed of a cell, an input gate, an output gate, and a forget gate. The cell remembers values over arbitrary time intervals, and the three gates regulate the flow of information into and out of the cell. In BiLSTM, one LSTM reads the relative location sequence forward, while a second LSTM reads it backward. The final two layers of hidden states are then concatenated, and the concatenation of these layers captures the speed and direction of users. To avoid overfitting, we perform both regular dropout and recurrent dropout.
FGLP adopts BiLSTM for two reasons: (1) LSTM works well for sequence modeling; (2) BiLSTM augments data by using backward sequences in training. While backward sequences are not part of the dataset, they are real sequences that could occur. Third, unlike unidirectional LSTM which leads to a final internal state containing more information about the last points of a sequence (while the information about the first points is forgotten)~\cite{DeBrebisson:2015:ANN:3056172.3056178}, BiLSTM preserves the sequence information equally across the relative points.

\textbf{CNN.} 
FGLP uses CNN on the historic region occupancy matrices, associated with the sequences fed into BiLSTM, to capture spatial features such as user movement preferences or the likelier route followed by a user between two points. CNN can learn this type of information because the historic region occupancy matrices contain information reflecting both occupancy frequency (explicit) and movement trajectory (implicit). Our CNN consists of batch normalization, convolution, max pooling, RELU activation, and dropout. With help from convolution and pooling, CNN is able to capture local connectivity and shift invariant. In our model, local connectivity can be the direction to which a user prefers to turn at a given intersection. The road characteristics are shift-invariant because  the road network in a city usually follows the same urban design, and is similar in different areas.

\textbf{Fusion.} 
FGLP fuses the output layers from BiLSTM and CNN by concatenation, which allows for different-length outputs from BiLSTM and CNN. Then the concatenated output is fed into fully connected densenets, and the final output is computed by {\em softmax} activation, as shown in equation~\ref{eq:softmax}, where $\mathbf{k}$ corresponds to the $\mathbf{k_{th}}$ grid-cell, n = M $\times$ M is the total number of grid-cells, $\mathbf{\hat{y}_k}$ is the $\mathbf{k_{th}}$ element of the output $\mathbf{\hat{Y}}$, and $\mathbf{\phi_k}$ is the $\mathbf{k_{th}}$ element of the final hidden layer before activation. The dense layers can gradually extract features of our desired length, and softmax converts them into probabilities. The output $\mathbf{\hat{Y}}$ contains the predicted probabilities of the future user location in each grid cell. 

\begin{equation}
  \label{eq:softmax}
  \mathbf{{\hat{y}_k} = \frac{\exp(\phi_k)}{\sum^{n}_{i=1} \exp(\phi_i)}}
\end{equation}

\begin{equation}
\label{eq:loss_function}
\mathbf{w^*} = \arg\min_{\mathbf{w}}{-\frac{1}{n}\sum_{i=1}^{n}{\mathbf{y_i} \cdot \log(\mathbf{\hat{y}_i})}}
\end{equation} 

FGLP uses cross entropy loss for optimization, which is a standard function in multi-class classification. Therefore, the model learns the parameter $\mathbf{w}$ by minimizing the cross entropy loss measurement (equation~\ref{eq:loss_function}). $\mathbf{y_i}$ is the $\mathbf{i_{th}}$ element of the ground truth $\mathbf{Y}$, where the grid cell of the user's future location is set to 1, and all others are 0.
}
\section{Dataset and Data Preprocessing}
\label{sec:data}
{\fontfamily{ptm}\selectfont
This section describes the two real-world public datasets that we use in our evaluation, as well as the data preprocessing to extract the inputs expected by our system.

\subsection{Dataset Description}

We use two datasets: Open PFLOW~\cite{KASHIYAMA2017249} and Geolife~\cite{4511454}. Most experiments will use Open PFLOW, which is much larger. Geolife is used to demonstrate model reusability.
Open PFLOW contains GPS trajectories that cover the Greater Tokyo area. It includes GPS data for 617,040 users, sampled every minute. The dataset covers typical movement patterns of people in the metropolitan area for one day, and it includes five transportation modes: stay, walk, vehicle, train, and bicycle. We select the data labeled ``walk'' and ``bycyle'', because  their lower speeds and ability to stop whenever they want are expected to enable more applications of predictions. 
Geolife~\cite{4511454} contains GPS trajectory data for 182 users over a five-year period. From this dataset, we selected the users (73) who labeled their trajectories with walking. 


\subsection{Data Preprocessing}





We choose a 200m $\times$ 200m region for both historic region occupancy matrices input and grid output, assuming a user can walk up to 100 meters in a minute. Let us recall that the region is centered at the current location of the user. For each experiment, we divide the region based on the accuracy we want to obtain. For example, we can divide the region into 100 cells of size 20m $\times$ 20m, and predict the location in one of these cells. The historic occupancy matrix is computed once and saved on mobile devices. The historic region occupancy matrix, centered at the current location, is extracted in real time from the overall occupancy matrix.

For the fixed-length sequences of relative points, we choose a length of 9 (i.e., 9 locations, sampled one minute apart). Any walking session less than 9 minutes is dropped from the input. The sequence length should contain enough information to capture the direction and speed of movement, but it should not be too long, delaying the time when the first prediction can happen. We selected $length=9$ as a good trade-off. In addition, most people in our datasets walk fewer than 15 minutes at once, so our length is practical from this point of view. 
To achieve quicker convergence, the relative points are scaled between -1 and 1. 

Both datasets contain a large number of standing-still datapoints. We consider an input sequence to be standing-still if the user does not move at all in 9 minutes (the sequence length). Even though there is a ``stay'' transportation mode, about 50\% of the walk data is standing-still. 
Our experiments presented in Section~\ref{sec:exp} are without standing-still data. However, we also ran experiments with standing-still data (not included in the paper for the sake of brevity). These experiments proved the imbalance introduced by standing-still data is not a problem for large amounts of data.

An accurate location prediction model requires large amounts of data. As we increase the region size to predict further into the future 
or decrease the grid cell size for higher accuracy, 
the number of grid cells increases quadratically. Therefore, the number of cells with insufficient samples increases correspondingly, and the prediction accuracy for both datasets may suffer. Furthermore, Geolife has fewer data samples and, thus, we do not expect to obtain good accuracy over this dataset. Nevertheless, we keep this dataset to test if our model can be reused by leveraging transfer learning.
 


}
\section{Experimental Evaluation}
\label{sec:exp}
{\fontfamily{ptm}\selectfont
Our evaluation has three goals. First, we assess the model's performance without FL. 
This allows a fair comparison between our model and the baseline models, which do not include privacy protection techniques such as FL. 
Second, we measure the performance of the federated solution of FGLP. 
Third, we test the feasibility of running FGLP on phones.

\subsection{Model Performance Without FL}

\subsubsection{Implementation}
FGLP is implemented using the Keras library. The output space of LSTM has dimensionality of 512 in each direction. Regular and recurrent dropout rates are set to 0.2 in BiLSTM. Two blocks of CNN are used sequentially with output space of 256 and 512, respectively. Both blocks include a convolutional layer with a $3 \times 3$ convolution window, a stride of 1, RELU activation, maximum pooling layer with size $2 \times 2$, batch normalization, and a dropout rate of 0.2. The outputs of BiLSTM and CNN are concatenated and fed into four densenet layers with output spaces of 1024, 512, 256, and 128, respectively. A dropout of 0.2 is added between the four dense layers. The final output is fitted by a densenet layer with softmax activation. We utilize the ADAM optimizer with learning rate of 0.001. The loss function is chosen as categorical cross entropy, which is commonly-used for multi-class classification. 

\subsubsection{Metrics}

We choose three metrics for prediction performance: cross entropy loss, categorical accuracy, and weighted F1 score (i.e., the weighted average of F1 scores of each class divided by the number of samples in each class). A high F1 score indicates that both precision and recall are high. 

\subsubsection{Baseline models}

We choose to compare FGLP with three baseline models: BiLSTM~\cite{DeBrebisson:2015:ANN:3056172.3056178}, CNN~\cite{8758930}, and HO. 
We also compare with a simple statistical prediction model: Highest Occupancy Model (HO). In HO, the location in the next minute is predicted as the most occupied grid cell in the historic region occupancy matrix. If there are multiple grid-cells with the same highest occupancy, HO will randomly select one of them.

To adapt the baseline models to be comparable with FGLP, we use the same input as FGLP, which are fixed-length sequences of relative points and historic region occupancy matrices. The dimensions of the hidden states and window sizes are chosen to be same as in FGLP, and eventually fed into the same output layers. In this way, the baselines can also demonstrate the contributions of each individual component of FGLP, and the benefits of fusing them.
We considered and tested two other state-of-the-art models for comparison, ST-RNN~\cite{10.5555/3015812.3015841} and DeepMove ~\cite{10.1145/3178876.3186058}, but we do not present the results due to their inferior performance compared to the performance of the other baselines shown in Table~\ref{compare}.

\subsubsection{Experimental Settings}
The experiments are conducted on a Ubuntu Linux cluster
(CPU: Intel(R) Xeon(R) CPU E5-2680 v4 @ 2.40GHz with 512GB memory, GPU: 4 NVIDIA P100-SXM2 with 64GB total memory), and the training and testing run with GPU acceleration. The ratio of training, validation, and testing datasets is 4:1:1.  We choose a batch size of 256. Early stopping is used to stop the training earlier if the accuracy has not improved in the last 10 epochs for the validation dataset. The users are simulated in this system by replaying their location traces.

\subsubsection{Results} 
Unless specified otherwise, the experiments are for pedestrian mobility, using Open PFLOW, and predicting one minute ahead for 20m $\times$ 20m grid-cells. As discussed in Section~\ref{sec:intro}, such short-term prediction can have significant benefits in real life. 



\textbf{Comparison with baseline models.}
Table~\ref{compare} shows the prediction metrics for FGLP without FL and the baseline models. FGLP outperforms the baselines in both loss and accuracy. The weighted F1 score indicates that both precision and recall are high in FGLP. We also notice that the CNN model performs better than the BiLSTM model, but individually none of them achieves good performance. Our model design manages to capture the essential information for the problem, and thus leads to substantially better performance.


\begin{table}
\centering
\caption{Performance of FGLP w/o FL and baselines}
\resizebox{0.32\textwidth}{!}{%
\begin{tabular}{|l|l|l|l|}
\hline
Model           & \multicolumn{3}{c|}{Metrics}              \\ \cline{2-4} 
                & Loss                      & Accuracy  & Weighted F1        \\ \hline
HO             & NA               & 0.329   & 0.382       \\ \hline
BiLSTM          & 2.859               & 0.222  &  0.188     \\ \hline
CNN             & 0.900               & 0.501   & 0.469       \\ \hline
FGLP            &  \textbf{0.157}               &  \textbf{0.955}   &  \textbf{0.955}       \\ \hline
\end{tabular}}
\label{compare}
\end{table}

\textbf{Performance vs. spatial scale.}
Figure~\ref{precision} shows how FGLP's accuracy varies with spatial scale (i.e., grid-cell size). Even though the accuracy decreases as the size of the cell decreases, FGLP can still achieve good performance (i.e., 87.8\% accuracy) for 5m $\times$ 5m cells. 

\begin{figure}
    \centering
    \resizebox{0.3\textwidth}{!}{%
\begin{tikzpicture}   
\centering   
\begin{axis}[
    xbar,
    enlarge x limits=0.2,
    bar width=0.4cm,
    height=3.7cm,
    width=8cm, 
    legend style={
        rotate=45,
        at={(1,1)},
        anchor=north east,
        legend columns=1,
        legend rows=2,
    },
    xlabel={Accuracy},
    ylabel={Grid Cell Size (meter)},
    symbolic y coords={
    20,
    10,
    5,
    },
    ytick = {{
    20,
    10,
    5,
    }},
    nodes near coords,
    nodes near coords align=horizontal,
    every node near coord/.append style={
        /pgf/number format/fixed zerofill,
        /pgf/number format/precision=3
    }
]
\addplot coordinates {(0.955,20) (0.894,10) (0.878,5)};
\end{axis} 
\end{tikzpicture}}
  \caption{Prediction accuracy as a function of grid-cell size}
  \label{precision}
\end{figure}
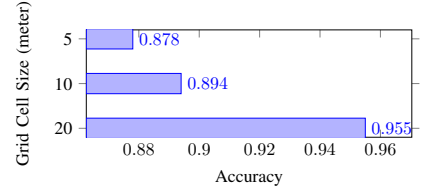

\textbf{Performance vs. time scale.}
Figure~\ref{min} shows how FGLP's accuracy varies over time (i.e., predict $x$ minutes ahead). The performance decreases over time, but we believe it is acceptable up to 5 minutes ahead. In addition to loss accumulation, there are two reasons for the decrease in accuracy over time. First, the user data is just for one day, and we cannot capture many recurring mobility patterns.
Second, pedestrian mobility is inherently difficult to predict. Nevertheless, short-term prediction can be beneficial to many applications. For example, adjusting bit rate video streaming based on 5G coverage prediction can allow buffering up enough video while coverage is still good to avoid quality degradation due to predicted poor coverage in the next few minutes. 

\begin{figure}
    \centering
    \resizebox{0.3\textwidth}{!}{%
\begin{tikzpicture}   
\centering   
\begin{axis}[
    xbar,
    enlarge x limits=0.2,
    bar width=0.4cm,
    height=4.5cm,
    width=8cm, 
    legend style={
        rotate=45,
        at={(1,1)},
        anchor=north east,
        legend columns=1,
        legend rows=2,
    },
    xlabel={Accuracy},
    ylabel={Time Window (min)},
    symbolic y coords={
    1st,
    2nd,
    3rd,
    4th,
    5th,
    },
    ytick = {{
    1st,
    2nd,
    3rd,
    4th,
    5th,
    }},
    nodes near coords,
    nodes near coords align=horizontal,
    every node near coord/.append style={
        /pgf/number format/fixed zerofill,
        /pgf/number format/precision=3
    }
]
\addplot coordinates { 
(0.9554606,1st)
(0.792224,2nd)
(0.78711265,3rd)
(0.70942986,4th)
(0.6675506,5th)
};
\end{axis} 
\end{tikzpicture}}
  \caption{Prediction accuracy as a function of time windows}
  \label{min}
\end{figure}
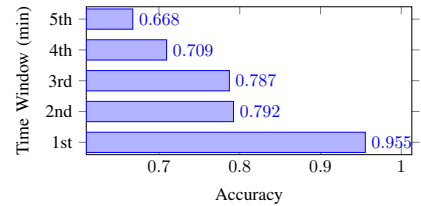

\textbf{Model Reusability.}~\label{reusability}
A model is reusable when it can be used directly or in conjunction with transfer learning (TL) on another dataset. Currently, this property exists mainly in image recognition and natural language processing because few models can satisfy it. 

We demonstrate that our model pre-trained on the pedestrian data from Open PFLOW works for bicyclist data from the same dataset, even without TL.  We also demonstrate that TL works well in conjunction with FGLP when applied to the Geolife dataset. We believe that the main reason for FGLP's reusability is its abstract data representation, which makes it less location-specific.

\begin{table}[t!]
\centering
\caption{FGLP w/o FL pre-trained on pedestrian data and used to predict on bicycling data w/o TL}
\resizebox{0.33\textwidth}{!}{%
\begin{tabular}{|l|l|l|l|}
\hline
Model           & \multicolumn{3}{c|}{Metrics}                                  \\ \cline{2-4} 
                & Loss  & Accuracy  & Weighted F1          \\ \hline
Walking         & 0.157                       & 0.955   & 0.955         \\ \hline
Bicycling       & 0.955                     & 0.853  & 0.849   \\ \hline

\end{tabular}}
\label{bike}
\end{table} 

Table~\ref{bike} shows the performance obtained when testing the pre-trained FGLP (i.e., trained only with pedestrian data) on bicycling data. For comparison, we also show the accuracy when testing on pedestrian data. Because bicycling speed is approximately four times walking speed, the sequence input is scaled down four times and the matrix input is scaled up four times to match the magnitude of pedestrian data. The results show good performance, with an accuracy of 85.3\%, which demonstrates the model's reusability.

Table~\ref{TL} shows the performance of FGLP over the Geolife dataset in two cases: a model trained directly on Geolife, and a model trained with TL from Open PFLOW to Geolife. The results demonstrate the reusability of FGLP because the TL model from Open PFLOW to Geolife achieves 12.2\% higher accuracy than the model trained on the Geolife dataset alone. This result is surprising, but it is explained by the fact that Open PFLOW is a much larger dataset than Geolife. 
We also observe, as expected, that training on Geolife leads to low accuracy. While the accuracy of the model pre-trained on Open PFLOW is significantly better, it is still not good in absolute terms. The reason is that the two datasets cover very different road networks. There are two types of urban planning for road networks, either as grid or circles, and these two types are quite different. Therefore, a model trained on one type cannot work very well on the other type. 

\begin{table}[t!]
\centering
\caption{FGLP performance on Geolife dataset:  Geolife alone vs. TL from Open PFLOW to Geolife}
\resizebox{0.45\textwidth}{!}{%
\begin{tabular}{|l|l|l|l|}
\hline
Model           & \multicolumn{3}{c|}{Metrics}                          \\ \cline{2-4} 
                & Loss  & Accuracy & Weighted F1        \\ \hline
Geolife alone         & 2.291                       & 0.400 & 0.390             \\ \hline
\shortstack{ TL from Open
PFLOW to Geolife} & 2.190                       & 0.448 & 0.441 \\ \hline

\end{tabular}}
\label{TL}
\end{table} 
}

\subsection{Model Performance with FL}
\label{sec:federated}
\begin{table}[t!]
\caption{FGLP with FL performance}
\resizebox{0.45\textwidth}{!}{%
\begin{tabular}{|l|l|l|l|}
\hline
Training method      & \multicolumn{3}{c|}{Metrics}     \\  \cline{2-4} 
            & Loss & Accuracy  & Weighted F1                \\ \hline
FGLP w/ FL without data augmentation       & 3.016           & 0.664       & 0.631   \\ \hline
FGLP w/o FL on augmentation dataset alone   & 2.652    &  0.792  &  0.815      \\ \hline
\textbf{FGLP w/ FL with data augmentation}     & \textbf{2.432}    & \textbf{0.842} & \textbf{0.853}\\ \hline
\end{tabular}}
\label{aug}
\end{table}

{\fontfamily{ptm}\selectfont
\subsubsection{Implementation and Settings}

FGLP with FL is implemented using TensorFlow Federated (TFF). For simplicity, we used the default federated averaging algorithm in TFF~\cite{DBLP:journals/corr/abs-1902-01046}, instead of more sophisticated averaging mechanisms that may help FL's accuracy~\cite{wang2020federated}. 
The same cluster is used for the experiments but we could not use GPU acceleration because current version of TFF (0.9.0) is not optimized for GPUs.

To simulate user participation in one FL round, we randomly sample 320 users every round for training. Before the training, we set aside some random users' data for testing, and these users are not selected for training. Because the number of samples per user is very limited (from 1 to 150, with an average of 60 in Open PFLOW), the batch size is set to 32. The number of users per round and the batch size are set experimentally for the best performance. All the other model settings are the same as in FGLP without FL. 
\\
\subsubsection{Results}

Table~\ref{aug} shows the performance of FGLP with FL. This experiment uses the training with data augmentation described in Section~\ref{sec:augment}.
During training, for each user selected in every round, we randomly pick 100 samples from the augmentation dataset, which are concatenated with the user dataset.
The best performance is achieved in the 4th round while training 20 epochs per user. FGLP with FL using data augmentation improves the accuracy by 20\% over the original FGLP with FL. 
These results demonstrates that our data augmentation mechanism can effectively improve the performance of the FGFL with FL model.

\subsection{Model Benchmarks on Smart Phones}
We implemented FGLP benchmark apps in Tensorflow and DL4J, and tested the feasibility of running FGLP on several smart phones for both training and inference in terms of latency, memory consumption, and battery consumption.  The maximum RAM usage of FGLP is less than 200MB. The Google Pixel 3XL, a mid-range phone in 2020, can make a prediction in 46.41ms, and train 200 samples in 15min. With a full battery, the Google Pixel 3XL can execute 12.9 million predictions, or 203 rounds of training. 
The results show the system works well in real-life. More details are omitted due to space constraints. 
}
\section{Conclusion and Future Work}
\label{sec:conclusion}
{\fontfamily{ptm}\selectfont
This paper proposed FGLP, a novel system for fine-grained location prediction that protects user privacy. The main novelties of FGLP are its prediction model, its abstract data representation, and its privacy-preserving learning framework. Our experiments demonstrated good prediction accuracy, model reusability, and system feasibility on smart phones. While FGLP is designed for predictions on smart phones, its model can also be used by network/service providers in their data centers. In the future, FGLP could be incorporated directly into smart phone OSs to improve system and app performance using location prediction. In addition, its GPS-based location prediction can be fused with location prediction done by wireless network providers based on 5G signal fingerprinting techniques to further improve the end-user's experience in real-world applications such as augmented reality, mobile gaming, and video streaming.

}
\balance


\bibliographystyle{plain}
\bibliography{sample-base}

\end{document}